**Pediatric ECG foundation model via adult-to-pediatric fine-tuning: downstream tasks and clinical evaluation**

Yu Cao[1,#], Haiyan Xu[2,#], Xiaohan Li[3], Jian Yang[1], Yuqing Feng[1], Wei Wang[4*], Haomin Li[1,3*]

1. Clinical Data Center, Children’s Hospital, Zhejiang University School of Medicine, National Clinical Research Center for Children and Adolescent's Health and Diseases, China
2. ECG Department, Children’s Hospital, Zhejiang University School of Medicine, National Clinical Research Center for Children and Adolescent's Health and Diseases, China
3. The College of Biomedical Engineering and Instrument Science, Zhejiang University, China
4. Department of Cardiology, Children’s Hospital, Zhejiang University School of Medicine, National Clinical Research Center for Children and Adolescent's Health and Diseases, China

# These authors contributed equally to this work

*Corresponding authors: hmli@zju.edu.cn; wangw_hz930@zju.edu.cn

**Abstract**

Pediatric electrocardiogram (ECG) interpretation is complicated by age-dependent electrophysiological changes that limit direct transfer of adult AI models. We asked whether an adult ECG foundation model could be adapted to pediatric ECGs. Using 339,563 12-lead ECGs from 238,189 children, we compared direct application, pediatric fine-tuning, and training from scratch under an identical architecture. Direct transfer was poor (mean AUC 0.8256). Fine-tuning reached 0.9695, above de novo training (0.9661), and attained 0.9079 using only 1% of training data. On 10,999 external ECGs, fine-tuning achieved a mean AUC of 0.9385 across 30 labels. It generalized to age regression (MAE 1.12 years) and congenital heart disease classification (AUC 0.9562). Across 17 clinicians and 100 cases, diagnostic accuracy rose from 56.8% to 65.4% ($p < 0.001$) and did not decline when AI erred. Adult ECG foundation models require pediatric adaptation; fine-tuning offers a data-efficient route to scalable pediatric ECG AI.

## Introduction

Electrocardiography (ECG) is a widely used non-invasive and economical diagnostic test that plays a crucial role in assessing cardiac health. While interpreting adult ECGs follows relatively standardized criteria, pediatric ECG interpretation is more complex, primarily because the physiological structure and electrophysiological characteristics of a child's heart undergo significant changes during growth and development[1]. Heart rate, PR and QRS intervals, QRS axis, precordial R/S progression, voltage amplitudes, and T-wave polarity all vary substantially by age, particularly during neonatal transition and infancy[2]. Contemporary pediatric reference studies have established age- and sex-specific normal limits rather than a single set of adult-derived thresholds[3,4]. These developmental changes create substantial overlap between physiological variation and findings that would be considered abnormal in adults. Consequently, adult ECG normal values, diagnostic criteria and decision support tools cannot be directly applied to children, posing unique challenges for automated pediatric ECG diagnosis. At the same time, experts with the ability to interpret pediatric ECG are scarce[5].

Currently, large-scale neural networks have achieved cardiologist-level arrhythmia classification, automated multi-label interpretation of 12-lead ECGs, detection of left ventricular systolic dysfunction, and identification of latent atrial fibrillation signatures during sinus rhythm[6–9]. These advances have established the ECG as a data-rich biomarker capable of encoding clinically relevant information beyond conventional visual interpretation[10]. Despite significant advancements in artificial intelligence (AI) for ECG analysis, most existing models trained on large-scale adult ECG data, face severe challenges in pediatric ECG applications[11]. Due to fundamental physiological developmental differences between children and adults, adult models often fail to accurately capture the unique physiological variations and pathological features present in pediatric ECGs, leading to significantly low accuracy and poor generalizability in pediatric diagnosis[12]. This adult-centric model design not only risks missed diagnoses or misdiagnoses of pediatric conditions but could also pose potential health risks to child patients. The scarcity of pediatric ECG datasets, the complexity of annotations, and the variability in physiological characteristics across different age groups severely limit the application of traditional AI algorithms in pediatrics. Developing AI-ECG models specifically for children not only requires overcoming the obstacle of insufficient data but also demands a deep understanding and effective integration of children's unique physiological developmental patterns.

In recent years, the emergence of foundation models has provided a new opportunity to address these challenges by learning reusable representations from large-scale data and adapting them to new tasks with comparatively limited task-specific supervision[13,14]. Relevant attempts have also been carried out in the field of ECG-AI: ECGFounder was pretrained on more than 10 million recordings across 150 diagnostic labels and was designed for both direct inference and downstream fine-tuning[15], while ECG-FM used large-scale self-supervised pretraining to produce transferable ECG representations for multiple clinical tasks[16]. More recent work has further emphasized that the clinical value of ECG foundation models depends not only on in-domain discrimination but also on transferability across tasks, populations, institutions, and acquisition settings[17]. As a routine, inexpensive, ubiquitous signal, ECG embeddings are emerging as opportunistic biomarkers, yielding unrequested risk data from existing recordings. Applying the concept of foundation models to pediatric ECGs holds the promise of building general diagnostic systems capable of adapting to the physiological characteristics of children at different developmental stages.

In this study, we investigate whether representations learned from a large-scale adult-centric ECG foundation model can be effectively transferred to pediatric populations through targeted fine-tuning on pediatric data, and subsequently serve as a reusable foundation for broader pediatric cardiovascular tasks. We hypothesize that while adult pretraining captures generic waveform and rhythm features, pediatric domain adaptation is essential to align these representations with age-dependent electrophysiological variations and diagnostic label distributions. To test this hypothesis, we conduct a systematic evaluation on 339,563 pediatric ECGs (0–18 years) and make three principal contributions. First, we perform a rigorously controlled comparison of three training paradigms—direct application of an adult pretrained model, pediatric fine-tuning, and training from scratch—using identical network architecture, input format, and evaluation metrics to isolate the effect of transfer learning. Second, we quantify the sample efficiency of pediatric fine-tuning across a spectrum of data availability (1% to 100% of the training set), demonstrating that high diagnostic performance can be achieved with substantially less labeled data than required for de novo training. Third, we move beyond conventional multi-label classification to evaluate the transferred representations on clinically meaningful downstream tasks, including ECG-based age regression and congenital heart disease screening, and further assess cross-institutional generalizability via an independent external dataset. Additionally, we explore the clinical usability of the fine-tuned model through clinical evaluations augmented by explainable visualizations, providing preliminary insights into AI-assisted pediatric ECG interpretation. Collectively, these contributions establish a data-efficient and scalable pathway for adapting adult ECG foundation models to pediatric care, while providing a multi-task evaluation framework that bridges artificial intelligence and pediatric digital medicine.

## Results

### Multi-model Diagnostic Comparison and External Validation

Of the 45 pediatric diagnostic labels evaluated, the adult pretrained model supported 33; the remaining 12 (e.g., right ventricular high voltage, peaked P wave) were absent from its output space and are marked as not applicable in Supplemental Table S2. On these 33 shared labels, the pretrained model yielded a mean AUC of 0.8256 (95% CI: 0.8140–0.8373), indicating limited zero-shot generalizability. For a matched comparison on the same 33-label subset, the fine-tuned and retrained models achieved mean AUCs of 0.9662 and 0.9638, respectively. Across all 45 labels, the fine-tuned model attained a mean AUC of 0.9695 (95% CI: 0.9659–0.9726) and the retrained model 0.9661 (95% CI: 0.9625–0.9695). Paired bootstrap analysis revealed a modest yet statistically significant improvement for the fine-tuned model ($\Delta$AUC = 0.0033, 95% CI: 0.0012–0.0056). Overall results are summarized in Figure 1a and Supplemental Table S2.

External validation was performed to assess cross-institutional robustness. As shown in Figure 1a and Supplemental Table S3, the fine-tuned model retained the highest performance on the Zhengzhou University pediatric ECG dataset, achieving a mean AUC of 0.9385 across 30 diagnostic labels. AUC values exceeded 0.90 for 23 of 30 labels, indicating robust discrimination for most diagnostic categories. The slight performance decline on the external validation set could be attributed to two primary factors. First, variations in ECG data sources (e.g., device vendors or acquisition protocols) introduced domain shift. More importantly, discrepancies existed in the pediatric ECG diagnostic criteria applied by experts from the two institutions. In China, national diagnostic standards for pediatric ECG remain lacking. Consequently, borderline findings such as mild T-wave changes may meet the reporting threshold at one institution but not at the other, reducing apparent agreement for these categories.

The progressive sample efficiency experiment revealed a clear data scaling effect during pediatric fine-tuning. As illustrated in Figure 1b, the fine-tuned model attained an AUC of 0.9079 using merely 1% of the pediatric training data. Performance improved progressively and peaked when the entire pediatric training set was utilized. Notably, the fine-tuned model nearly matched the performance of the retrained model with just 25% of the training data and surpassed it beyond the 50% threshold. Consequently, the full-data fine-tuned model was adopted as the final model for all subsequent downstream analyses.

Figure 1c presents a label-wise AUROC comparison across models, highlighting classification performance on 21 labels that were shared between the internal and external test sets and contained more than 10 cases in the internal test set. The most substantial improvements were predominantly observed in ST change (STC), first degree AV block (1AVB), long QT-interval (LONGQT) and Nodal escape beat (NEB). Children exhibit a higher prevalence of physiological mimics, such as STC from normal repolarization, 1AVB with shorter PR limits. This physiological overlap with pathology makes adult-derived thresholds inadequate and adult-centric model reduced diagnostic accuracy in pediatric ECG analysis.

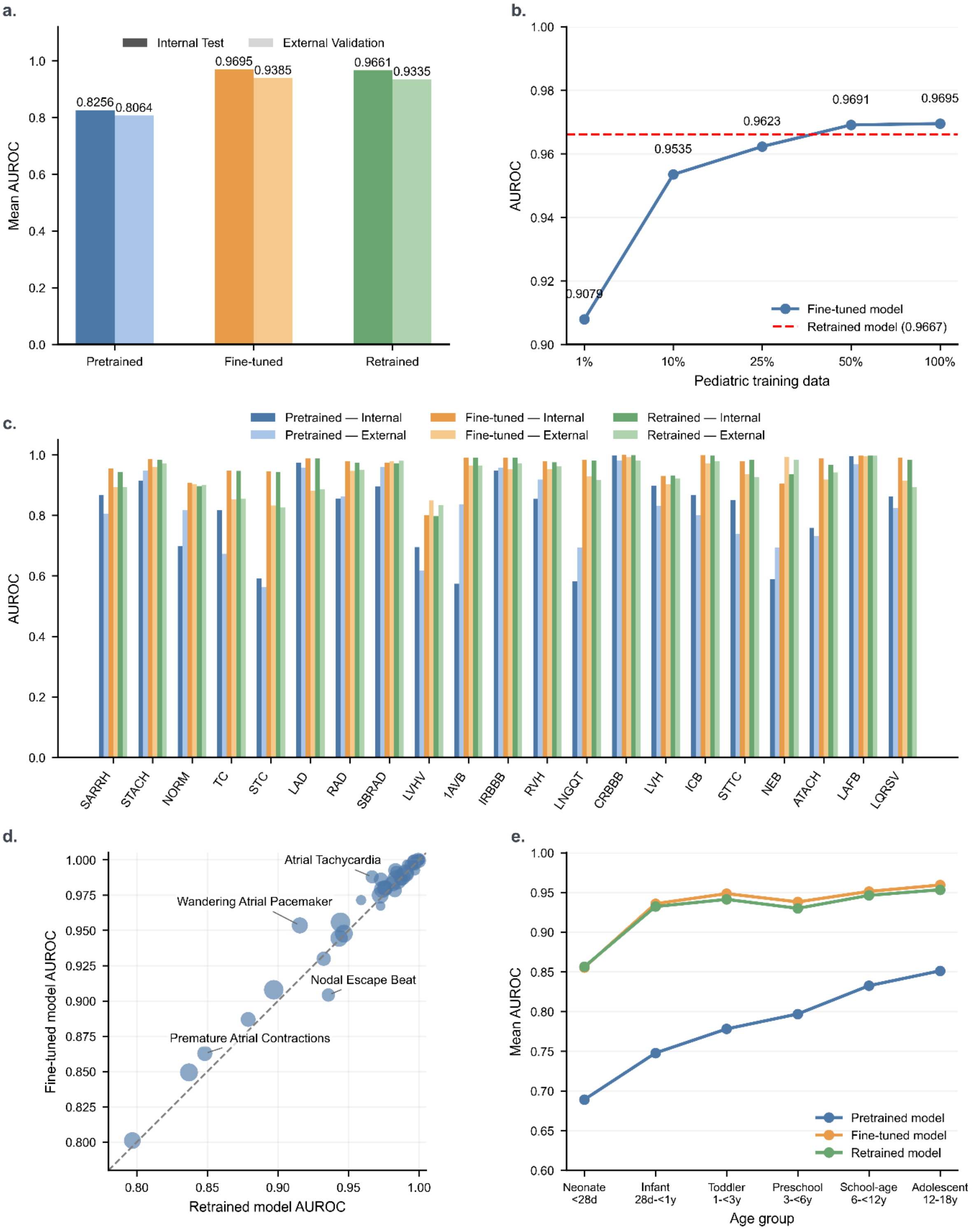

**Figure 1. Multi-model Diagnostic Comparison and External Validation. a.** Mean AUROC of the pretrained, fine-tuned, and retrained models' performance on internal test dataset and external validation dataset. **b.** Sample efficiency analysis of progressive pediatric fine-tuning. **c.** Label-wise AUROC comparison across shared ECG labels with more than 10 ECG recordings. **d.** Paired label-wise AUC comparison between the fine-tuned model and the retrained model. **e.** Age-stratified AUC comparison across pediatric developmental stages.

The pediatric fine-tuned model marginally outperformed the retrained model overall, suggesting that low-level waveform features learned from large-scale adult ECG pretraining remain transferable but require pediatric domain adaptation to capture pediatric-specific electrophysiological characteristics. Beyond aggregate performance metrics, we conducted a label-wise paired comparison between the fine-tuned and retrained models. As shown in Figure 1d, across 45 diagnostic labels, the fine-tuned model achieved a higher AUC for 33 labels, a lower AUC for 7 labels, and identical AUCs for 5 labels. These findings suggest that adult ECG pretraining not only enhances average performance but also confers subtle, label-level advantages for a broad set of pediatric ECG diagnoses following pediatric domain adaptation. Overall, the tight clustering of data points around the identity line underscores the robustness of both pediatric training paradigms, whereas the positive bias in the delta plot positions fine-tuning as the more data-efficient strategy.

Age-stratified analysis further demonstrated that pediatric adaptation improved diagnostic performance across all developmental stages. As shown in Figure 1e, both the fine-tuned and retrained models consistently outperformed the adult pretrained model in every age group, with the largest absolute performance gap observed in younger children. The fine-tuned model exhibited a slight yet consistent advantage over the retrained model across most age strata, supporting the value of adult pretraining followed by pediatric domain adaptation. All three models attained their lowest AUC values in the neonatal group (detailed performance were shown in Supplemental Table S4), which had the least training data and the greatest physiological complexity, whereas performance improved progressively from infancy to adolescence.

**Downstream Age Regression Task**

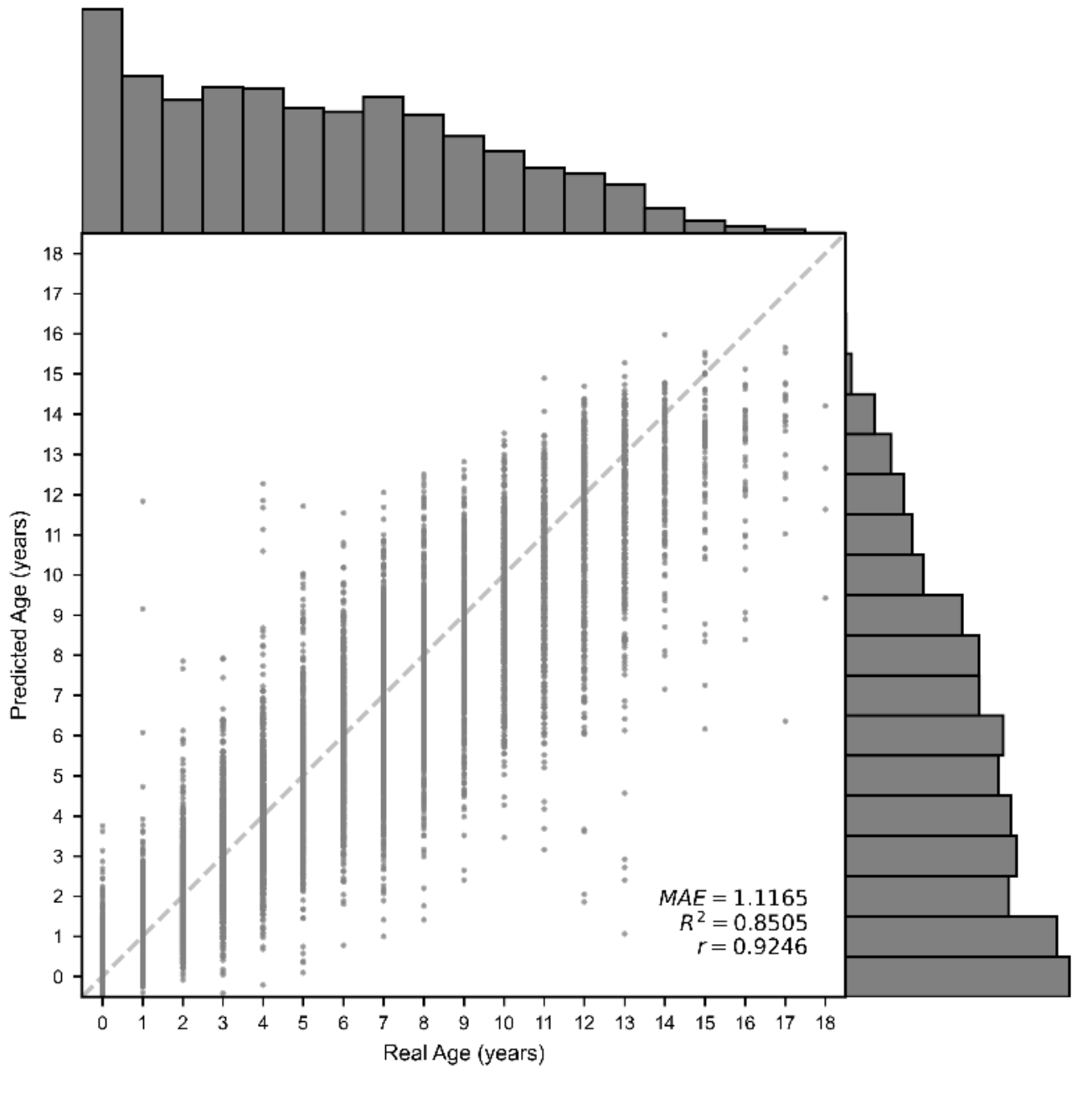

**Figure 2. The predicted age by pediatric ECG foundation model and real age in test dataset.**

In the age regression task, the pediatric fine-tuned model achieved a mean absolute error (MAE) of 1.1165 years, corresponding to $R^2$= 0.8505, and Pearson r = 0.9246. The pediatric fine-tuned model also slightly outperformed the retrained model (MAE = 1.1331 years). As a reference, in a previous adult ECG-age study, the MAE was 8.38 years[18]. This finding suggests that the model encoded latent developmental electrophysiological information in pediatric ECGs.

The scatter plot in Figure 2 shows a strong correlation between predicted and chronological age, with predictions generally aligned along the identity line. However, mild overestimation in younger children and mild underestimation in older adolescents reflect a regression-toward-the-mean effect commonly observed in biological age modeling, attributable to the inherent imbalance in the data distribution. In external validation, the model achieved an MAE of 1.6602 years. This performance discrepancy likely stems from the external validation cohort consisting of hospitalized children with greater disease severity. Prior work has shown that disease status in children can bias age predictions[19]. These results further demonstrate that the features learned by the model exhibit cross-center generalization capability, potentially aiding in the assessment of biological age from pediatric ECGs and the analysis of development-related ECG characteristics.

**Downstream Congenital Heart Disease Classification Task**

The downstream congenital heart disease (CHD) classification task demonstrated consistently high discriminative performance. As shown in Table 1, across CHD subtypes, the retrained model achieved a mean AUC of 0.9433, while the fine-tuned model achieved a mean AUC of 0.9562, both of which exceeded the CHDdECG baseline of 0.8872. Compared with traditional task-specific deep neural networks such as CHDdECG, downstream models fine-tuned from the pediatric ECG foundation model achieved an average improvement of 5.1% in recall, while demonstrating a more substantial gain of 11.6% in specificity. This suggests that foundation models learn deep representations that task-specific networks fail to capture.

**Table 1. Overall performance of downstream congenital heart disease classification tasks**.

| CHD | Positive | Retrained Model | | | Fine-tuned Model | | | CHDdECG Model | | |
|---|---|---|---|---|---|---|---|---|---|---|
| subtype | Count | AUC | Recall | Specificity | AUC | Recall | Specificity | AUC | Recall | Specificity |
| VSD | 1627 | 0.9761 | **0.9205** | 0.9225 | **0.9798** | 0.9177 | 0.9225 | 0.9473 | 0.8371 | **0.9286** |
| ASD | 1631 | 0.9290 | 0.8016 | 0.9258 | **0.9386** | **0.8036** | **0.9267** | 0.8948 | 0.7829 | 0.8585 |
| PDA | 759 | 0.9196 | 0.8129 | **0.8654** | **0.9322** | **0.8259** | 0.8626 | 0.8433 | 0.7143 | 0.8386 |
| TOF | 118 | 0.9972 | 0.9658 | **1.0000** | **0.9995** | 0.9316 | **1.0000** | 0.9819 | **0.9746** | 0.9153 |
| TGA | 28 | 0.9180 | 0.8148 | **0.9333** | **0.9556** | **0.9259** | 0.8667 | 0.8320 | 0.8606 | 0.5532 |
| CoA | 23 | 0.9197 | **0.8846** | 0.7826 | **0.9314** | 0.8462 | **0.9565** | 0.8238 | 0.7778 | 0.7432 |

In the external validation dataset, the fine-tuned model continued to demonstrate robust discriminative performance for congenital heart disease (CHD) labels. Specifically, the AUC for VSD (n=1,062) was 0.9242, while that for ASD (n=1,606) reached 0.9464, both indicating strong discriminatory power. This suggests that the model exhibited good generalization ability for common CHDs across multi-center data. Notably, the performance achieved for ASD was superior to that on the internal test set, primarily attributable to the typically larger defect size of ASD observed in hospitalized children.

**ECG Explainability Analysis and Clinical Evaluation**

Finally, 10 full-time electrocardiographers with routine ECG reporting duties and 7 pediatric cardiologists participated in the clinical evaluation. To assist clinicians in interpreting the predictions of the black-box model, Grad-CAM visualizations were integrated into the evaluation platform. Representative Grad-CAM cases are provided in the supplementary material. During the evaluation, predicted labels were presented alongside their corresponding Grad-CAM visualizations.

Overall, the diagnostic accuracy of clinicians improved significantly from 56.80% to 65.40% ($p < 0.001$) as shown in Figure 3a. As shown in Figure 3b, without AI assistance, professional electrocardiographers demonstrated significantly higher diagnostic accuracy than general cardiologists (64.70% vs. 45.40%). With AI assistance, both groups showed marked improvement. Specifically, the accuracy of electrocardiographers increased by 5.2 percentage points ($p = 0.008$), while that of cardiologists improved by 13.5 percentage points ($p = 0.016$). Notably, although the average performance of general cardiologists in interpreting pediatric ECGs with AI assistance remained lower than that of specialized electrocardiographers, the performance gap narrowed, and the two groups of clinicians showed no statistically significant difference after using AI.

Figure 3c illustrates the diagnostic accuracy across different types of ECG. Clinicians showed significantly higher accuracy in identifying normal and conduction abnormalities, but performed poor on labels such as ventricular hypertrophy & high voltage and ST-T & Q-wave abnormality. Although diagnostic performance varied across different ECG categories, AI assistance enhanced the diagnostic capabilities of clinicians in every category. This is especially evident in the category where clinical staff performed worst (ST-T and Q-wave abnormality). Figure 3d displays the ranking of diagnostic accuracy changes for each specific ECG label before and after AI assistance, ordered from largest to smallest. Among all 29 labels, 21 showed improved diagnostic accuracy, while the remaining 8 showed no change, and no labels deteriorated. Notably, accuracy increased by more than 20 percentage points for several labels, including peaked P waves (PPW), premature atrial contractions (PAC), wandering atrial pacemaker (WAP), third-degree AV block (3AVB), intraventricular conduction block (ICB), ST changes (STC), and sinus arrhythmia (SARRH). Diagnosing these pediatric ECGs presents significant challenges, primarily due to the extensive overlap between physiological variants and pathological conditions. For instance, wandering atrial pacemaker (WAP) and sinus arrhythmia (SARRH) are often benign variants in children and can be easily confused with pathological states, while intraventricular conduction block (ICB) and peaked P waves (PPW) frequently yield false positives due to physiological factors such as a thin chest wall and right ventricular dominance. Furthermore, ST changes (STC) exhibit high non-specificity, making isolated ECG-based diagnosis difficult.

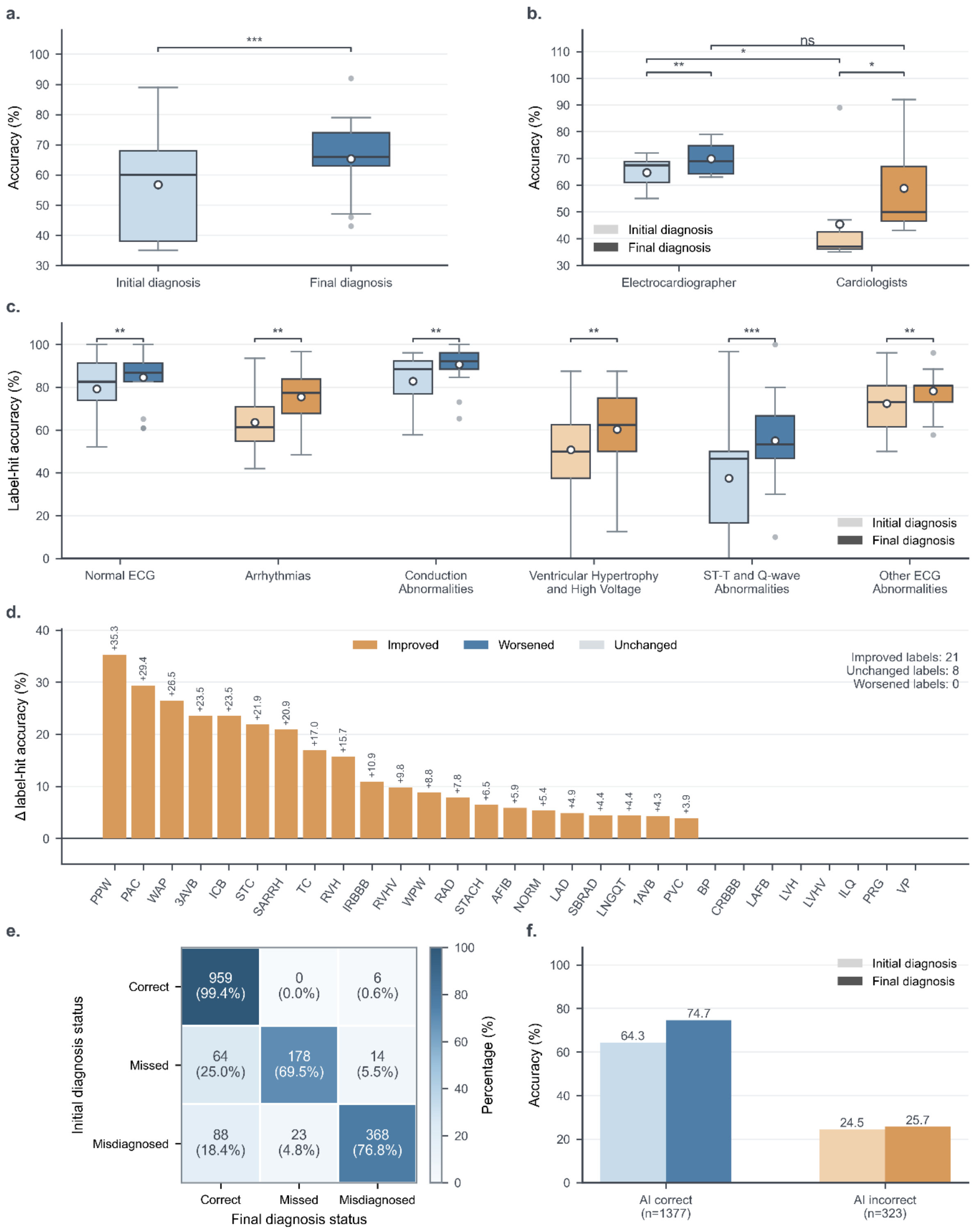

**Figure 3. Clinical Evaluation. a.** Diagnostic accuracy of all clinicians before and after AI assistance. **b.** Diagnostic accuracy of professional electrocardiographers and cardiologists before and after AI assistance. **c.** Changes in diagnostic accuracy across different ECG groups before and after AI assistance. **d.** Difference in diagnostic accuracy for specific ECG labels before and after AI assistance. **e.** State matrix of correct, missed, and misdiagnosed diagnoses before and after AI assistance. **f.** Impact of cases with correct versus incorrect AI results on clinical diagnosis.

The foremost concern regarding the clinical deployment of ECG-AI is whether the model might mislead clinicians, and to what extent. Figure 3e illustrates the changes in diagnostic status before and after AI integration. Among cases where clinicians initially made a correct diagnosis, only 0.6% were misled into an incorrect diagnosis after using AI. For diagnoses with initially omitted labels, AI assistance helped retrieve 25.0% of these omissions. Furthermore, in cases where clinicians originally assigned incorrect labels, 18.4% were corrected to the correct labels with AI support. As shown in Figure 3f, when cases were stratified by AI prediction correctness, clinicians generally achieved higher accuracy on cases that AI predicted correctly. With AI assistance, diagnostic accuracy significantly improved for cases with correct AI predictions (74.7% vs. 64.3%, $p < 0.001$). Importantly, even in cases where AI predictions were incorrect, no significant misleading effect was observed; instead, the actual diagnostic accuracy slightly increased (25.7% vs. 24.5%). In other words, AI errors did not lead to catastrophic outcomes when interpreted by professional clinicians. Based on self-reported diagnostic confidence (on a 1–5 scale) during the clinical evaluation, the average confidence score after AI assistance was 4.14, representing a 0.34-point increase from 3.80 before AI use ($p = 0.011$).

## Discussion

This study demonstrates that although the adult-centric ECG foundation model was validated on large-scale datasets, its direct application to pediatric populations—particularly younger children—yielded suboptimal performance. Although adult and pediatric ECGs share fundamental waveform components (e.g., P waves, QRS complexes, and T waves), normal reference ranges and abnormal manifestations in children are highly dependent on developmental stage, especially regarding conduction intervals, repolarization features, QRS axis, and heart rate.

Relative to training from scratch, pediatric fine-tuning following adult pretraining yielded enhanced overall performance with considerably less training data. This indicates that while large-scale adult pretraining encodes generic ECG features (e.g., waveform and rhythm patterns), pediatric fine-tuning is crucial for mitigating age-related physiological discrepancies and aligning with pediatric-specific label distributions. Given that annotated pediatric ECG data are expensive to acquire and public datasets are limited, this transfer-learning paradigm offers substantial practical utility.

Age stratification analysis demonstrated that neonates had the lowest overall diagnostic performance. Several physiological and data-related factors may account for this relatively lower AUC: (1) Neonatal ECGs diverge most substantially from adult ECGs due to rapid postnatal cardiovascular transition, including elevated heart rate, right ventricular dominance, rightward electrical axis, immature conduction, and dynamic repolarization. These developmental characteristics can mimic or overlap with pathological patterns, blurring the line between normal variation and disease. (2) The neonatal period is extremely brief, and ECG morphology evolves rapidly within days of birth; thus, a single age category may contain marked physiological heterogeneity. (3) Neonatal ECG recording is particularly prone to motion artifacts, challenges related to small body size, and variability in lead placement, all of which elevate signal noise. (4) Compared to older children, neonatal diagnostic labels are scarcer and more imbalanced, restricting the availability of positive training examples. Together, these factors explain the persistent difficulty of neonatal ECG interpretation despite pediatric fine-tuning, highlighting the necessity for neonatal-specific calibration, expanded datasets, and age-stratified modeling in future studies.

The downstream age regression task provides additional evidence supporting the potential of the pediatric ECG foundation model as a powerful tool for pediatric ECG representation. Specifically, the high correlation coefficient and low mean absolute error suggest that the model's latent features are associated with both diagnostic categories and continuous electrophysiological changes throughout development. This inherent developmental awareness accounts for the model's strong performance in pediatric cohorts even in the absence of age-specific input features. Nevertheless, pediatric cardiologists usually rely on age information to inform their interpretation of ECG data, highlighting a gap between model learning and clinical reasoning.

The CHD downstream tasks further demonstrate that pediatric ECG representations contain clinically meaningful signals associated with structural heart disease. While ECG remains a non-invasive adjunct rather than the definitive diagnostic modality for CHD, the robust AUC values achieved for VSD, TOF, PDA, and CoA imply that the model discerns subtle electrophysiological alterations secondary to hemodynamic shifts, ventricular volume/pressure overload, conduction delays, or axis deviations. When benchmarked against the SOTA model CHDdECG, our foundation-model-based framework exhibited enhanced discriminative capability, underscoring that pretraining on large-scale pediatric ECG data fosters richer, more transferable representations than narrow task-specific detectors. This benefit was most pronounced for complex or heterogeneous defects (e.g., TGA, PDA, CoA), where pathological ECG patterns are typically understated and diffuse across various waveform features[20]. These results highlight the promise of the pediatric ECG foundation model as a scalable, non-invasive screening adjunct for early CHD detection and risk stratification.

The present work is highly complementary to the recent study by Mayourian et al.[12], in which a task-specific deep neural network trained on 583,134 pediatric ECGs from Boston Children's Hospital demonstrated expert-level performance for binary abnormality detection (AUROC 0.94), Wolff-Parkinson-White syndrome (AUROC 0.99), and prolonged QTc (AUROC 0.96). Employing multi-label classification, our model achieved competitive performance across 45 diagnostic labels, delivering superior results for WPW syndrome (AUROC 0.99) and prolonged QTc (AUROC 0.98). A key strength of their investigation was the incorporation of blinded expert readjudication, showing that experts were more likely to concur with the AI model than with the original cardiologist reads. Moreover, our own clinical evaluation indicated that the explainability of visualized AI explanations can aid clinical experts in more precise pediatric ECG analysis. Nevertheless, their task-specific model targeted a narrow range of outcomes and did not evaluate transferability to wider applications. Conversely, our foundation model approach specifically explored the adaptation of adult-pretrained representations to pediatric ECGs, attaining comparable multi-label performance across 45 diagnostic labels while exhibiting greater data efficiency via transfer learning. Furthermore, our study encompassed external cross-institutional validation, progressive sample efficiency analysis, age-stratified evaluation, and downstream tasks such as age regression and congenital heart disease screening. Collectively, these two studies illustrate complementary strategies for pediatric ECG AI: task-specific models tailored for focused clinical deployment versus foundation models offering transferable representations for diverse pediatric cardiovascular applications.

The clinical evaluation of this study confirmed two principal conclusions. First, both professional electrocardiographers and pediatric cardiologists improved their ECG interpretation performance using this pediatric ECG foundation model, with pediatric cardiologists deriving even greater benefits. Second, the prior concern that ECG-AI might lead to blind adherence among clinicians proved unfounded in our evaluation. Even when the test set deliberately included a higher proportion of cases with incorrect AI predictions than the baseline assessment, the use of AI in these scenarios still slightly enhanced clinician performance. These findings provide supportive evidence for the safe and widespread implementation of ECG-AI in pediatric care.

Several limitations should be acknowledged. First, while external validation was conducted on an independent pediatric ECG dataset from Chinese children, the model's training data originated from a single institution. This necessitates further multicenter, cross-regional, and cross-device validation to confirm generalizability. Second, the limited sample sizes for low-prevalence (rare) diagnostic labels may lead to unstable F1 scores, particularly for uncommon abnormalities. Third, the CHD downstream tasks used age-matched 1:1 case–control sampling, which differs from real-world screening prevalence; prospective validation in a consecutive cohort is required to establish PPV and NPV. Lastly, the current framework mainly utilized ECG waveforms and structured labels, without comprehensively incorporating demographic and clinical information (e.g., age, sex, and medical history). Theoretically, integrating these multimodal features would further improve model performance and robustness.

In summary, this study demonstrated that direct transfer of an adult ECG foundation model to pediatric populations yielded limited performance, whereas pediatric fine-tuning substantially improved diagnostic accuracy and slightly outperformed training from scratch using pediatric data alone. Downstream age regression results further indicated that the model captures developmental electrophysiological representations inherent in pediatric ECGs. Additionally, congenital heart disease classification tasks expanded the clinical scope of the model, demonstrating that pediatric ECG foundation representations can be effectively transferred to clinically meaningful structural heart disease screening. Clinical evaluation demonstrated that using the pediatric ECG foundation model generally improves clinicians' ability to interpret pediatric ECGs, and moreover, even when the AI produced erroneous predictions, their clinical performance did not deteriorate. Collectively, these findings motivate future work extending pediatric ECG-AI to longitudinal outcome prediction, including mortality and long-term cardiovascular events. Overall, the pediatric ECG foundation model offers a viable pathway for building generalizable and scalable pediatric ECG diagnostic systems, while establishing a methodological foundation for future pediatric cardiovascular screening, risk stratification, and AI-assisted diagnosis.

## Methods

The model comparison experiments evaluated three strategies: (i) the **Pretrained Model**, which involved directly applying an adult ECG foundation model to pediatric ECG data; (ii) the **Fine-tuned Model**, initialized from the adult ECG foundation model and fine-tuned on pediatric training data; and (iii) the **Retrained Model**, which utilized the same architecture but was trained from scratch exclusively on pediatric training data.

We conducted a multi-model comparison to assess the transferability of an adult ECG foundation model to pediatric ECG tasks. To ensure fair comparison and isolate the impact of training strategies, all models utilized identical input formats, architectures, evaluation metrics, and an independent test set, thus preventing performance variations from being confounded by differences in test data.

### Dataset and Pre-processing

This retrospective study was approved by the Institutional Review Board (IRB) of the Children's Hospital, Zhejiang University School of Medicine (Approval No. 2023-IRB-0108-P-01; Date: May 3, 2023). In accordance with the committee's regulations and the Declaration of Helsinki, the requirement for informed consent was waived for the use of de-identified retrospective data. The experimental data were sourced from a large-scale, standardized pediatric ECG dataset, comprising 339,563 standard 12-lead ECG recordings from 238,189 children (aged 0–18 years) who underwent ECG examinations at the same institution between January 2019 and June 2023. Males accounted for 59.9% of the cohort, while females comprised 40.1%. All ECGs underwent two-stage review by two different qualified ECG diagnostic cardiac electrophysiologists in practice reporting routine. The interpretation and classification of pediatric ECGs adhered to the Zhejiang Pediatric Routine Electrocardiogram Diagnostic Standards (First Edition), established by the Non-invasive Electrocardiology Group of the Zhejiang Medical Association. The original Chinese ECG statements were mapped to annotation labels following the SCP-ECG v0.4 standard[21]. If no corresponding standard label existed, new labels were defined according to relevant principles. The distribution of 69 ECG annotation labels is detailed in Supplemental Table S1. Each recording has a duration of 10 seconds and was sampled at 500 Hz.

Given the significant age-related changes in children's cardiac physiological parameters, the dataset used in this study spans all developmental stages—from neonates, infants, and toddlers to school-aged children and adolescents—ensuring adequate representation for each age group. This enables the model to capture ECG features across diverse age ranges. Figure 4 illustrates the age and gender distribution of the dataset, with age groups defined according to the work of Bratincsák et al.[3].

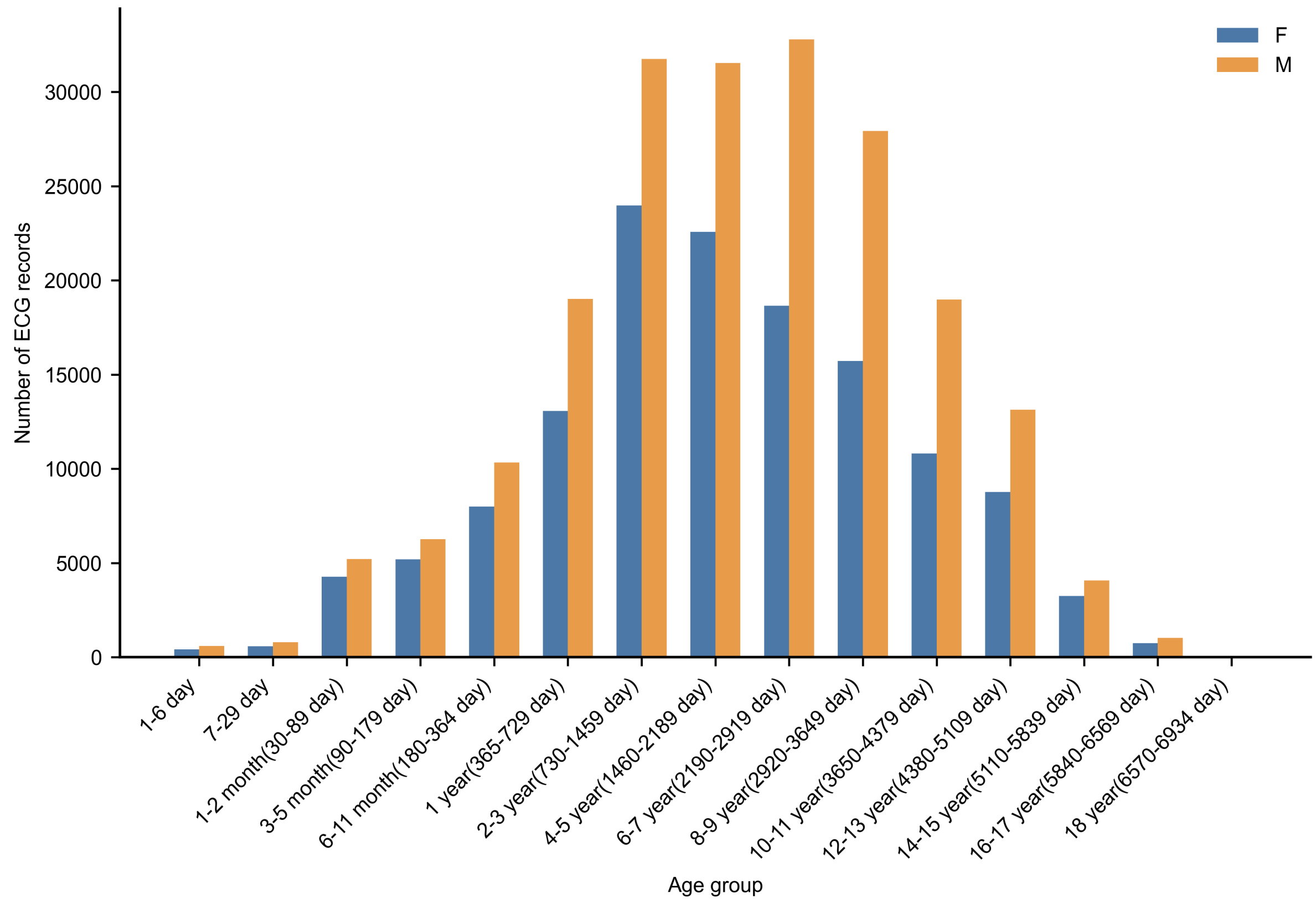

**Figure 4. Age and sex distribution of the pediatric ECG dataset used for this study.**

Signal enhancement and quality control are particularly critical in pediatric ECG analysis. Therefore, high-pass, low-pass, and notch filtering techniques were applied to suppress baseline drift, high-frequency noise, and power frequency interference[4]. Furthermore, incomplete or missing lead data were handled using a zero-padding approach, thereby ensuring data integrity and maximizing the utilization of available information.

For independent evaluation, a test set of 10,000 ECG records was randomly selected. Subsequently, 500 ECG recordings from 238 children with duplicate entries in the training set were excluded, resulting in a final internal test set comprising 9,500 subject-level independent ECG test data. Only 45 labels were show up in the test dataset. The remaining 329,563 records were then randomly partitioned into training and validation sets following an 80:20 split. Specifically, the training set was used for model training and fine-tuning, while the validation set was employed for early stopping, and hyperparameter optimization.

**Model Architecture and Training Strategy**

All experiments employed the RegNet-style one-dimensional convolutional architecture adopted by the ECGfounder model[15]. This architecture utilizes a stage-wise design to capture both temporal waveform features and spatial inter-lead relationships, making it well-suited for standard 12-lead ECG analysis. The model architecture is illustrated in Supplemental Figure S1.

All three training strategies shared the same architecture and evaluation pipeline. (1) The **Pretrained Model** evaluated the zero-shot generalizability of adult ECG representations when applied to pediatric ECGs. It should be noted that the pretrained model had 150 labels but did not cover all pediatric ECG labels defined in this study (e.g., RVHV, which is common in children). (2) The **Fine-tuned Model** examined the efficacy of adapting adult pretrained weights to pediatric data. To investigate the volume of pediatric ECG data required for effective adaptation, we conducted a progressive sample efficiency study. Specifically, the pediatric training dataset was randomly subsampled at five proportions—1%, 10%, 25%, 50%, and 100%—yielding 3,302, 32,956, 82,391, 164,782, and 329,563 ECG records, respectively, while maintaining the original label distribution. The model was fine-tuned sequentially on each subset under identical optimization settings. Performance on the held-out pediatric test set was evaluated after each stage to quantify the relationship between training data scale and model performance. (3) The **Retrained Model** assessed whether training from scratch exclusively on pediatric data could serve as a viable alternative to adult pretraining.

For pediatric fine-tuning, we initialized the model with the ECGFounder checkpoint pretrained on adult data. Architecturally, the model comprises seven sequential feature-extraction stages, with channel dimensions of 64, 160, 160, 400, 400, 1024, and 1024, and block counts of 2, 2, 2, 3, 3, 4, and 4, respectively. While all pretrained backbone parameters were retained, weights from the original dense classification head were discarded. Finally, the output layer was substituted with a newly initialized fully connected layer mapped to the pediatric diagnostic label space with 69 labels.

All models were trained and evaluated as multi-label classifiers. Training was conducted with a batch size of 32 and an initial learning rate of $1 \times 10^{-4}$. We employed the Adam optimizer with a weight decay of $1 \times 10^{-5}$, and optimized the multi-label classification task using binary cross-entropy with logits (BCEWithLogits) loss. For learning rate scheduling, a ReduceLROnPlateau strategy was adopted: the validation macro-AUROC was monitored, and the learning rate was multiplied by a factor of 0.1 if no improvement was observed for three consecutive epochs (patience = 3). Finally, optimal classification thresholds for individual diagnostic labels were determined on the validation set by maximizing Youden's J statistic.

To assess external generalizability, the three models were additionally evaluated on an independent external pediatric ECG dataset from Zhengzhou University[22]. This external test set comprised 10,999 standard 12-lead ECG recordings from children aged 0–14 years, covering 30 diagnostic labels that overlapped with our internal dataset.

We further conducted an age-stratified subgroup analysis to evaluate whether model performance remained consistent across pediatric developmental stages. The independent test set was categorized into six clinically meaningful age groups: neonates (<28 days), infants (28 days to <1 year), toddlers (1 to <3 years), preschool children (3 to <6 years), school-age children (6 to <12 years), and adolescents (12 to 18 years). For each age stratum, the mean macro-AUROC was calculated separately for the Pretrained, Fine-tuned, and Retrained Models.

**Downstream Tasks**

Age is a key reference parameter in pediatric ECG interpretation. To evaluate whether the pediatric ECG foundation model had learned general developmental representations beyond disease classification, we designed a downstream age regression task. This task predicted chronological age from ECG waveforms and assessed whether the model captured age-dependent electrophysiological characteristics. The same training/test dataset was used for this downstream task. The Mean Absolute Error (MAE) between predicted and actual ages was used as the evaluation metric.

Beyond the regression task, we evaluated clinically relevant downstream classification tasks for common pediatric congenital heart diseases (CHDs). A downstream classifier was trained to identify ventricular septal defect (VSD), atrial septal defect (ASD), tetralogy of Fallot (TOF), patent ductus arteriosus (PDA), transposition of the great arteries (TGA), and coarctation of the aorta (CoA) based on ECG-derived representations. This enabled us to assess whether the learned pediatric ECG features could support structural heart disease screening beyond standard ECG diagnostic labels. For each CHD subtype, an independent binary classification task was constructed. A total of 20,943 CHD ECG records were collected. For each subtype the positive cases were split into training, validation, and test sets at a 7:1:2 ratio. The same subject-level isolation was applied to the test set to ensure that only one ECG recording per CHD child was included in the test dataset. The whole dataset for each task comprised ECGs from patients with the target CHD and age-matched normal controls at a 1:1 ratio, ensuring a balanced distribution of positive and negative samples. Normal controls were randomly sampled from the healthy pediatric ECG population to match the number of disease cases. To establish a clinically relevant benchmark, we used the previously published CHDdECG model by Chen et al.[20] as a baseline comparator for these downstream tasks. Both pediatric ECG foundation model variants and CHDdECG were trained and evaluated using identical disease categories and metrics.

For each downstream task, we initialized the model with the best fine-tuned checkpoint and discarded the original classification head. Parameters from the dense classification layer were excluded, and a new task-specific fully connected layer (output dimension aligned with the target task) was added. Instead of linear probing, we performed full end-to-end fine-tuning, jointly optimizing all backbone parameters (convolutional and intermediate blocks) and the new classification head via back-propagation.

**Clinical Evaluation**

The model's performance does not directly correlate with clinical manifestations. With the model's assistance, clinical experts will formulate their own judgments based on its recommendations, which may include both valuable insights and potentially misleading information. To help users understand the black-box nature of ECG-based AI, we integrated Gradient-weighted Class Activation Mapping (Grad-CAM) visualizations alongside predictive labels to assist clinical interpretation[23].

We enrolled 10 full-time pediatric electrocardiographers from the ECG department and 7 pediatric cardiologists from the cardiology department from the Children's Hospital, Zhejiang University School of Medicine to evaluate a well-defined cohort of 100 ECG cases. These cases were selected based on pediatric ECG diagnostic labels to ensure maximum diversity in both diagnostic categories and patient ages. Furthermore, to assess the potential misleading impact of model errors, we deliberately included 19 cases with incorrect model predictions (see Supplemental Table S5 for details).

The evaluation comprised two distinct stages. In the first stage, experts reviewed the ECG signals and key measurement values to select diagnostic labels and provide confidence ratings (on a 1–5 scale, with 5 indicating the highest confidence). Upon submission, the system revealed the ECG foundation model's predicted labels and Grad-CAM-enhanced visualizations. Subsequently, clinicians were asked to confirm or revise their label selections and provide an updated confidence rating. The detailed clinical evaluation web pages were shown in supplemental material. To mitigate blind trust in AI outputs, all participating clinicians were explicitly told prior to the evaluation that AI predictions might be erroneous, as we had intentionally included cases where the model failed. Diagnostic accuracy was defined as exact label matching, where in multi-label cases, any omission or incorrect labeling was considered an error.

**Ethics Approval**

This retrospective study received approval from the Institutional Review Board/Ethics Committee of Children's Hospital of Zhejiang University School of Medicine (approval No. 2023-IRB-0108-P-01; May 3, 2023). Informed consent was waived per institutional guidelines for de-identified retrospective data. All procedures adhered to the Declaration of Helsinki principles.

**Data Availability**

The pediatric ECG dataset used in this study is available from the Children's Hospital, Zhejiang University School of Medicine upon reasonable request and subject to appropriate data use agreements. All data will be gradually made available at http://pecg.nbscn.org. The external validation dataset is publicly available at https://doi.org/10.6084/m9.figshare.27078763.

**Code Availability**

The code for model training and evaluation will be made available at https://github.com/Rain914/Pediatric-ECG-Foundation-Model upon publication.

**Author contributions**

Model training and testing: Y.C. and H.L.; ECG records acquisition: H.X., Y.C., X.L., W.W. and H.L.; Data statistics and analysis: Y.C., Y.F., J.Y., and H.L.; Website developing: Y.C., and H.L.; Clinical evaluation: Y.C., H.X., W.W. and H.L.; Manuscript preparation: Y.C., H.X., W.W. and H.L.; Critical comments and revision of the manuscript: all authors.

**Acknowledgements**

This work is supported by the Natural Science Foundation of China (No. 82572377) and the "Pioneer" and "Leading Goose" R&D Program of Zhejiang (No. 2026C01022).

**Competing Interests**

The authors declare no competing financial or non-financial interests.

**Figure legends:**

**Figure 1. Multi-model Diagnostic Comparison and External Validation. a.** Mean AUROC of the pretrained, fine-tuned, and retrained models' performance on internal test dataset and external validation dataset. **b.** Sample efficiency analysis of progressive pediatric fine-tuning. **c.** Label-wise AUROC comparison across shared ECG labels with more than 10 ECG recordings. **d.** Paired label-wise AUC comparison between the fine-tuned model and the retrained model. **e.** Age-stratified AUC comparison across pediatric developmental stages.

**Figure 2. The predicted age by pediatric ECG foundation model and real age in test dataset.**

**Figure 3. Clinical Evaluation. a.** Diagnostic accuracy of all clinicians before and after AI assistance. **b.** Diagnostic accuracy of professional electrocardiographers and cardiologists before and after AI assistance. **c.** Changes in diagnostic accuracy across different ECG groups before and after AI assistance. **d.** Difference in diagnostic accuracy for specific ECG labels before and after AI assistance. **e.** State matrix of correct, missed, and misdiagnosed diagnoses before and after AI assistance. **f.** Impact of cases with correct versus incorrect AI results on clinical diagnosis.

**Figure 4. Age and sex distribution of the pediatric ECG dataset used for this study.**